# A Genetic Algorithm Meta-Heuristic for a Generalized Quadratic Assignment Problem

Mojtaba A. Farahani [a], Alan McKendall [a*]

[a] Department of Industrial and Management Systems Engineering, Benjamin M. Statler College of Engineering and Mineral Resource, West Virginia University, Morgantown, WV 26506, United States

*Corresponding author: Alan.McKendall@mail.wvu.edu

## Abstract

The generalized quadratic assignment problem (GQAP) is one of the hardest problems to solve in the operations research area. The GQAP addressed in this work is defined as the task of minimizing the assignment and transportation costs of assigning a set of facilities to a set of locations. The facilities have different space requirements, and the locations have different space capacities. Multiple facilities can be assigned to each location if the space capacity is not violated. In this work, three instances of GQAP in different situations are presented. Then, a genetic algorithm is developed to solve the GQAP instances. Finally, the local neighborhood search with the steepest descend strategy is constructed and applied to the final solution obtained by the GA, and the final solution is compared with the best solution found by MPL/CPLEX software and reference papers. The results show that the developed GA heuristic is effective for solving the GQAP.

## Keywords

Generalized Quadratic Assignment Problem, Genetic Algorithms, Metaheuristics, Neighborhood Search Techniques

The generalized quadratic assignment problem (GQAP) is one of the hardest problems to solve in the operations research area. The GQAP addressed in this work is defined as the task of minimizing the assignment and transportation costs of assigning a set of facilities to a set of locations. The facilities have different space requirements, and the locations have different space capacities. Multiple facilities can be assigned to each location if the space capacity is not violated. In this work, three instances of GQAP in different situations are presented. Then, a genetic algorithm is developed to solve the GQAP instances. Finally, the local neighborhood search with the steepest descend strategy is constructed and applied to the final solution obtained by the GA, and the final solution is compared with the best solution found by MPL/CPLEX software and reference papers. The results show that the developed GA heuristic is effective for solving the GQAP. The main parts of this work are adopted from IENG 554 lecture notes, [1], and [2]. The supplementary files, MATLAB codes and problem files of this work are available at https://github.com/tamoraji/GA_for_GQAP.

# 1. Problem Definition & Mathematical Model

The generalized quadratic assignment problem (GQAP) assigns a set of machines (M) to a set of locations (N locations) where M > N such that more than one machine can be assigned to a location based on the machine's requirements and the capacities of the locations while minimizing the sum of the assignment and transportation costs [3].

## 1.1.　　　Non-linear Problem Definition

The number of units of materials transported between machine $i$ and $j$ ($f_{ij}$), the distances between locations $k$ and $l$ ($d_{kl}$), the space requirement of each machine $i$ ($r_i$), the capacity of each location k ($c_k$), the costs of assigning each machine i to each location k ($a_{ik}$), and the unit cost per distance unit of transporting materials between each pair of machines i at location k and machine j at location l ($c_{ijkl}$) are deterministic and known. The non-linear mathematical formulation of the problem will be as follows:

$x_{ik} = 1, if\ machine\ i\ is\ assigned\ to\ location\ k\ and, 0\ otherwise$

$$Min\ Z\ = \sum_{i=1}^{M}\sum_{k=1}^{N} a_{ik}\,x_{ik} + \sum_{i=1}^{M}\sum_{j=1}^{N}\sum_{k=1}^{N}\sum_{l=1}^{M} c_{ijkl}\,f_{ij}\,d_{kl}x_{ik}x_{jl} \qquad (1)$$

$$j \neq i \qquad l \neq k$$

$$Subject\ to\ \sum_{k=1}^{N} x_{ik} = 1,\ \ for\ i = 1,2,\ldots,M \qquad (2)$$

$$(3)$$

$$\sum_{i=1}^{M} r_i\,x_{ik} \leq c_k,\ \ for\ k = 1,2,\ldots,N$$

$$(4)$$

$x_{i,k} = 0\ or\ 1, \forall\ i,k$

Objective function (1) minimizes the sum of the assignment/installation and material handling costs. Constraint set (2) ensures that each machine is assigned to only one location. Constraint set (3) ensures that the space capacity of each location is not exceeded, and the restrictions on the decision variables are given in (4) [1].

## 1.2.    Linear Problem Definition

The objective function in previous section, contain a quadratic term. As a result, the mathematical formulation (1)– (4) is nonlinear and need a nonlinear programming model. As we know, the nonlinear programming techniques does not guarantee an optimal solution and we need to linearize the model to be able use linear programming techniques and reach to an optimal solution. The model is linearized by substituting $w_{ijkl}$ for the quadratic term $x_{ik}x_{jl}$. So, replacing objective function (1) with (1a) we will have:

$$Min\ Z\ = \sum_{i=1}^{M}\sum_{k=1}^{N} a_{ik}\,x_{ik} + \sum_{i=1}^{M}\sum_{j=1}^{N}\sum_{k=1}^{N}\sum_{l=1}^{M} c_{ijkl}\,f_{ij}\,d_{kl}w_{ijkl} \qquad (1a)$$

And adding new constraints (5) and (6):

$$(5)$$

$$x_{ik} + x_{jl} - 1 \leq w_{ijkl} \ \forall \, i, j \neq i \ and \ \forall \, k, l \neq k$$

$$w_{ijkl} \geq 0 \ \forall \, i, j \neq i \ and \ \forall \, k, l \neq k \tag{6}$$

The linearized model is a mixed integer linear programming model (MILP) for the GQAP. This model will be used in the next section to solve a small GQAP instance using MPL/CPLEX software.

### 1.3. MPL/CPLEX formulation

Figure.1. shows the MPL formulation that has been used in this project solve three different problem instances mathematically using linear programming techniques. M is the number of machines and N is the number of locations, respectively, and the DATA and transportation cost part in the formulation should be substituted based on each problem instance accordingly.

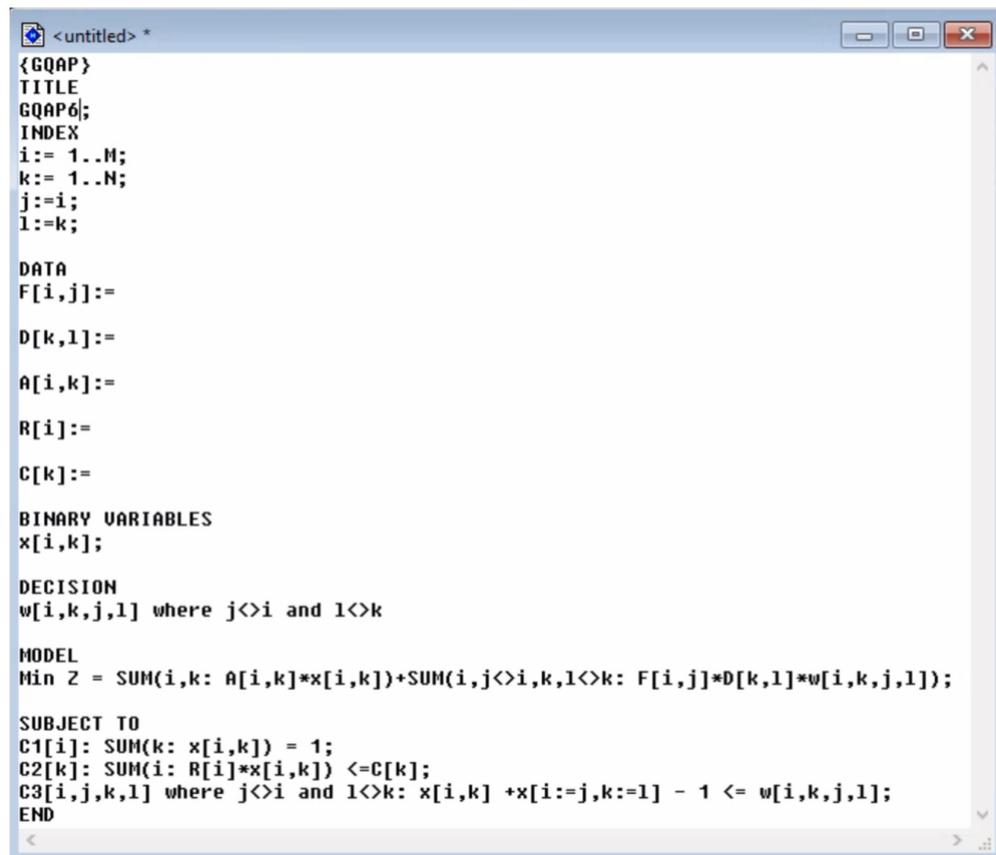

```
<untitled> *

{GQAP}
TITLE
GQAP6;
INDEX
i:= 1..N;
k:= 1..N;
j:=i;
l:=k;

DATA
F[i,j]:=

D[k,l]:=

A[i,k]:=

R[i]:=

C[k]:=

BINARY VARIABLES
x[i,k];

DECISION
w[i,k,j,l] where j<>i and l<>k

MODEL
Min Z = SUM(i,k: A[i,k]*x[i,k])+SUM(i,j<>i,k,l<>k: F[i,j]*D[k,l]*w[i,k,j,l]);

SUBJECT TO
C1[i]: SUM(k: x[i,k]) = 1;
C2[k]: SUM(i: R[i]*x[i,k]) <=C[k];
C3[i,j,k,l] where j<>i and l<>k: x[i,k] +x[i:=j,k:=l] - 1 <= w[i,k,j,l];
END
```

Figure.1: MPL formulation used in this project

## 2. Solving GQAP instances with LP and MPL/CPLEX results

Three different GQAP instances have been used in this project. a small instance, a medium instance, and a large instance. The small instance of a problem is the task of assigning six machines to four locations. The medium instance problem is the task of assigning 20 machines to 15 locations, and the large instance problem is the task of assigning 50 machines to 10 locations. The details of each problem instance are given in the project assignment and will not be discussed here. Here, we will present the results achieved by using the CPLEX solver using MPL/CPLEX software and the linearized mathematical formulation for the GQAP presented before. It should be noted that the software was run on the WVU virtual machine with 16 GB of RAM and an Intel Xeon Platinum 8272CL at 2.6 GHz.

For the small problem instance, the MILP has 370 constraints and 384 variables, which include 24 integers. The optimal solution was found by the software in under a second. The optimal objective function value for this problem is, and the optimal solution is: $x13 = x21 = x34 = x42 = x51 = x61 = 1$, and all other decision variables are zero. More specifically, machines 2, 5, and 6 are assigned to location 1. Machines 4, 1, and 3 are assigned to locations 2, 3, and 4, respectively. Figure 2 is the screenshot of the problem solution.

```
    MPL Modeling System   -   Copyright (c) 1988-2012, Maximal Software, Inc.
    ---------------------------------------------------------------------------

    MODEL STATISTICS

    Problem name:        GQAP6M4L (64-bit)

    Filename:            GQAP
    Date:                November 15, 2022
    Time:                11:17
    Parsing time:        0.01 sec

    Solver name:         CPLEX  (12.4.0.0)
    Objective value:     17165.0000000
    Integer nodes:       0
    Iterations:          149
    Solution time:       0.07 sec
    Result code:         101

    Constraints:         370
    Variables:           384
    Integers:            24
    Nonzeros:            1128
    Density:             0.8 %

    SOLUTION RESULT

      Optimal integer solution found

         MIN Z           =      17165.0000
```

Figure. 2: MPL/CLEX result for the small instance problem

For the medium problem instance, the MILP has 79835 Constraints and 80100 variables which include 300 integers. The software ran out of memory after about three hours and 18 minutes and could not reach to the optimal solution and objective function value. The best-found objective function value for this problem is $Z = 1714264$ . This value will be used as the benchmark for the best-found solution by MPL/CPLEX. Figure. 3 is the screenshot of the problem solution

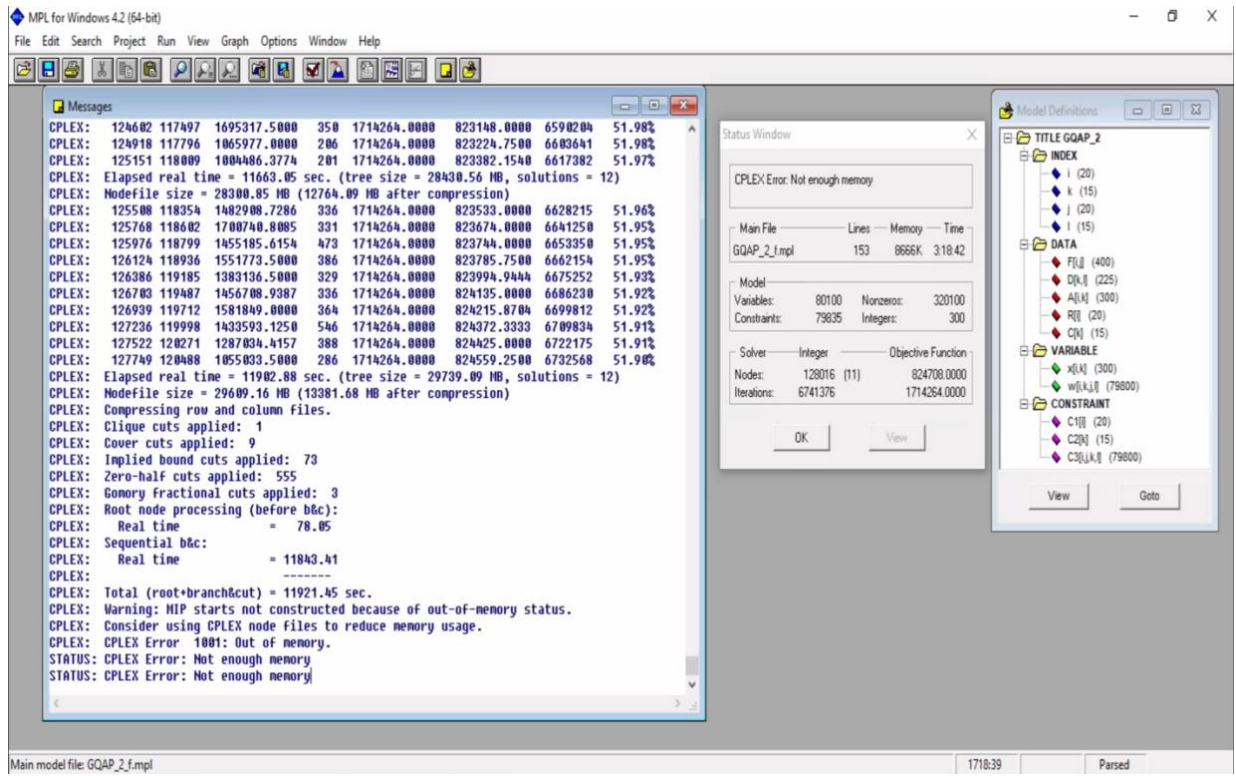

Figure. 3: MPL/CLEX result for the medium instance problem

For the large problem instance, the MILP has 220560 Constraints and 221000 variables which include 500 integers. The virtual machine logged out after about five hours and could not reach to the optimal solution and objective function value. The best-found objective function value for this problem is $Z = 12878101$. This value will be used as the benchmark for the best-found solution by MPL/CPLEX. Figure. 4 is the screenshot of the problem solution

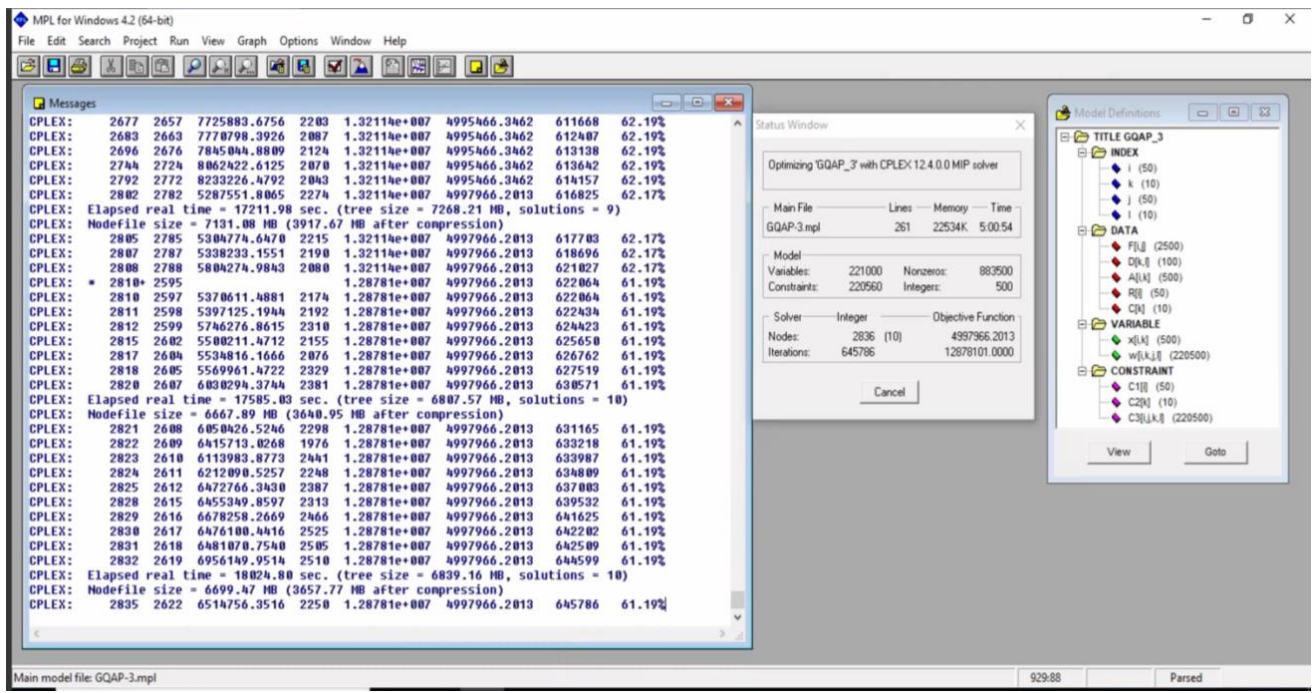

Figure. 4: MPL/CLEX result for the medium instance problem

# 3. Solving GQAP instances using Genetic Algorithm Metaheuristics

The general framework adopted in this project is presented as Figure 5. The detail of each step will be discussed in subsequent subsections.

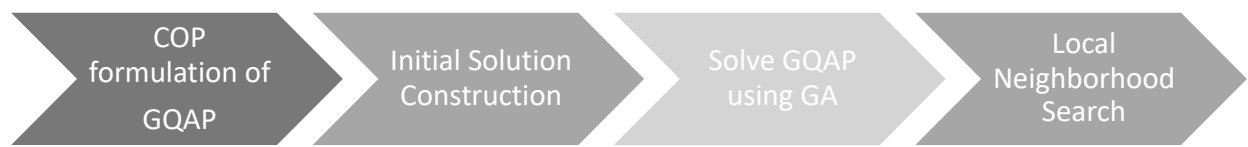

Figure. 5: The framework to address the GQAP

## 3.1. Solution representation

The mathematical model that was defined in Section 1 can only be used to solve small problems in a reasonable amount of time. Thus, heuristic and meta-heuristic algorithms are developed for the GQAP.

As it was shown in [1], it is much more efficient using the COP model, as opposed to the mathematical model, to solve the GQAP. The combinatorial optimization problem (COP) model and the solution representation for our problem are as follows:

$S = (S(1), S(2), \ldots, S(M))$ where S(i) = k means machine i is assigned to location k

$$Min\ TC(S) = \sum_{i=1}^{M} a_{is(i)} + \sum_{i=1}^{M} \sum_{\substack{1 \leq j \leq M \\ j \neq i}} c_{ijs(i)s(j)} f_{ij} d_{s(i)s(j)}$$

$$subject\ to \sum_{\forall i\ s.t.s(i)=k} r_i \leq C_k\ for\ k = 1, \ldots, N$$

For example, the optimal solution for the small problem instance presented before is represented as S = (3, 1, 4, 2, 1, 1). That is, s(1) = 3, s(2) = 1, s(3) = 4, s(4) = 2, s(5) = 1, and s(6) = 1. More specifically, machines 2, 5, and 6 are assigned to location 1, machine 4 to location 2, machine 1 to location 3, and machine 3 to location 4.

## 3.2.    Genetic Algorithm

A genetic algorithm (GA) is an intelligent probabilistic search algorithm that simulates the process of evolution by constructing a population of solutions and applying genetic operators (i.e., crossover and mutation) in each reproduction. Each solution in the population is evaluated according to the objective function and fitness of the solution. Highly fit solutions in the population are given opportunities to reproduce and generate offspring. New offspring solutions are generated, and unfit solutions in the population are replaced. This evaluation-selection-reproduction cycle is repeated until a satisfactory solution is found or a stopping criterion has been met [2]. In this project, I adopted the genetic algorithm proposed by Chu and Beasly [2] with a few minor modifications for GQAP. First, the general steps of the algorithm will be presented, and then each step will be discussed in more detail along with a numerical example from the small problem instance of the project. Figure 6 is the general GA algorithm that has been used in this project. The details of each step will be discussed subsequently. The MATLAB code snippet of each step and the results of the code for the small instance will be presented.

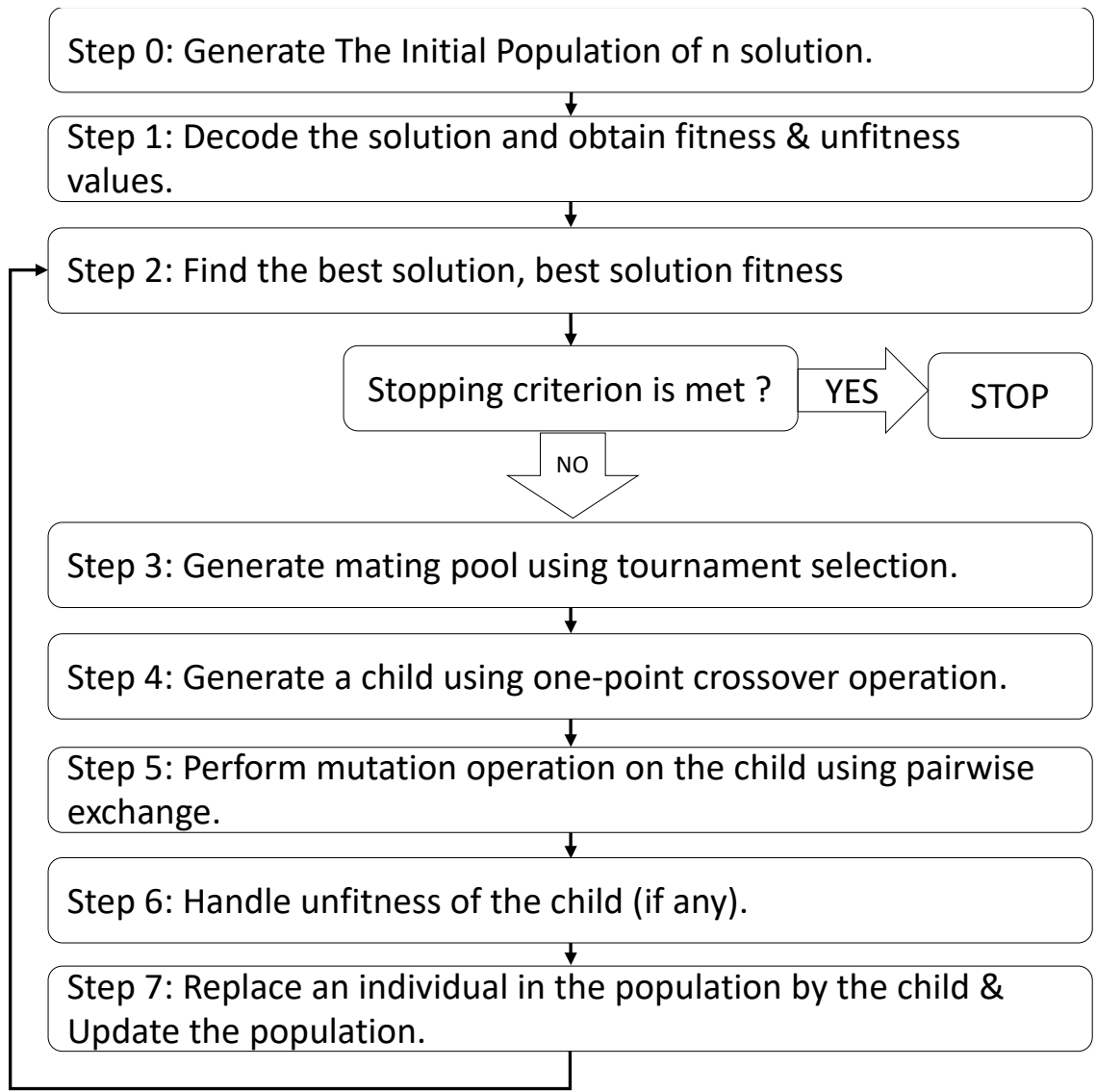

Figure. 6: The GA algorithm flowchart

**Step 0:** In this step, we generate an initial population of n randomly constructed solutions. Each of the initial solutions is generated by randomly assigning a machine to a random location. Note that the initial solutions may violate the capacity constraint and be infeasible. The number of chromosomes in the initial population is defined by the user.

**Step 1:** In this step, we calculate the fitness and unfitness values of each chromosome in the population. F, D, A, C, and R are the problem input data matrices. The notation is the same as the notation that was defined in the COP model. "costcalc" is the function to calculate the fitness value based on the defined COP model, and "unfitness_calc" is the function to calculate the unfitness of the solutions. The unfitness of a solution is a measure of infeasibility (in relative terms) as calculated by the formula in [2]. It should be noted that the unfitness value is equal to 0 if the solution is feasible. Figure 7 illustrates the code snippets to calculate these two values.

```
function [extra_cap,unfitness] = unfitness_calc(pop,C,R)
unfitness = zeros(size(pop,1),1);
extra_cap = zeros(size(pop,1),length(C));
for n = 1:size(pop,1)
    for i = 1:length(C)
        found = pop(n,:)==i;
        cap_ext = sum(R(found))-C(i);
        extra_cap(n,i) = cap_ext;
        unfitness(n) = unfitness(n) + max([cap_ext,0]);
    end
end
end

function cost = costcalc(pop,F,D,A)
cost = zeros(size(pop,1),1);
for n = 1:size(pop,1)
    for i = 1:size(pop,2)
        for j = 1:size(pop,2)
            if j == i
                continue
            end
            cost(n) = cost(n) + F(i,j)*D(pop(n,i),pop(n,j));
        end
    end
    for i = 1:size(pop,2)
        cost(n) = cost(n) + A(i,pop(n,i));
    end
end
end
```

Figure. 7. costcalc and unfitness_calc funtions

Figure. 8. is the code snippet for step 0 and step 1. In Figure 8, "*n_pop"* is the number of chromosomes in the population (e.g., 5), "N" is the number of locations, and "M" is the number of machines.

```
%% initialization

tic
%rng('default')
P0 = randi(N,n_pop,M); % Generate 'n_pop' random solutions
fitness_P0 = costcalc(P0,F, D,A);
[extra_cap_P0,unfitness_P0] = unfitness_calc(P0,C,R);
final_P0 = [P0 int16(fitness_P0) unfitness_P0];
disp(final_P0)
```

Figure. 8. step 0 & step 1: generate the initial population

Table 1 shows an example of the initial population constructed by the algorithm in steps 0 and 1.

Table 1. Initial population example for small instance problem

| Solution | | | | | | Fitness | Unfitness |
|---|---|---|---|---|---|---|---|
| 4 | 4 | 3 | 4 | 4 | 3 | 13030 | 400 |
| 2 | 2 | 1 | 1 | 2 | 3 | 18438 | 120 |
| 2 | 4 | 4 | 2 | 1 | 4 | 16340 | 190 |
| 3 | 2 | 3 | 1 | 3 | 2 | 20130 | 200 |
| 4 | 1 | 3 | 4 | 4 | 4 | 18785 | 280 |

**Step 2:** The next step is to find the population's best solution and determine its fitness. Since there can be a case where an unfit solution has a lower fitness value than a fit solution, the best solution is chosen only from fit solutions, and the solution that has the minimum fitness value is considered the best solution. If there is no solution with unfitness equal to zero in the population, the solution with the least amount of unfitness is chosen as the best solution. The fitness value for this solution is set to a large number (e.g., 999,999,999). Figure 9 shows the code snippet for this step.

```
%% Find the best_sol P0 & initial z_best

found = find(unfitness_P0==0);
if isempty(found)
    [~,ind] = min(unfitness_P0);
    best_sol = P0(ind,:);
    z_best = 9999999;
else
    fitted = final_P0(found,:);
    [~,ind] = min(fitted(:,M+1));
    best_sol = fitted(ind,1:M);
    z_best = fitted(ind,M+1);
end
```

Figure. 9. step 2: find the best solution in current population

**Stopping criterion:** The stopping criterion should be checked before moving on to the next step of the algorithm. We defined a parameter "*K_iter*" here, which is defined as the number of iterations that a new non-duplicate child is generated, but the best solution was not improved. The stopping criterion is for the algorithm to run for a predefined number of iterations (e.g., "maxiter = 100000") or for *K_iter to* reach a predefined value. (e.g., 'max_k=100'). The algorithm goes to step 3 if either of those two criteria is not met. If any of them are met, the algorithm stops and reports the best-found solution and the respected fitness value.

**Step 3:** Select two parent solutions for reproduction. the tournament selection scheme used for this step. Two individuals are chosen randomly from the population. The more fit individual is then allowed into the mating pool. To produce a child, two tournaments are held, each of which produces one parent. Note that the selection criteria do not involve the unfitness value of an individual [2]. Duplicate parents are also not acceptable in the mating pool. Figure 10. shows the code snippet and an example of it for the small problem instance.

```
% Generate the mating pool with the tournament scheme
M_pool = zeros(2,M);
choose = randperm(size(P0,1),2); %Choose 2 random solutions
if fitness_P0(choose(1)) <= fitness_P0(choose(2)) %compare the fitness values
    M_pool(1,:) = P0(choose(1),:);
else
    M_pool(1,:) = P0(choose(2),:);
end
ii=1;
while ii < 2
    choose = randperm(size(P0,1),2); %Choose 2 random solutions
    if fitness_P0(choose(1)) <= fitness_P0(choose(2)) %compare the fitness values
        if P0(choose(1),:) == M_pool(1,:)
            continue
        end
        M_pool(2,:) = P0(choose(1),:);
        ii = 2;
    else
        if P0(choose(2),:) == M_pool(1,:)
            continue
        end
        M_pool(2,:) = P0(choose(2),:);
        ii = 2;
    end
    if ~isempty(find(M_pool(2,:), 1))
        break
    end
end
```

```
new generation is:
     1      2      3      1      4      1   22273      10
     3      1      1      2      4      1   19373      10
     2      3      4      1      1      2   19610      20
     4      4      2      1      1      3   20400     110
     1      1      4      2      4      1   18878     110

Iteration Number:
     6

The mating pool is:
     3      1      1      2      4      1
     2      3      4      1      1      2
```

Figure. 10: step 3: generating mating pool from the population

In this example, second and third solution randomly chosen for the mating pool.

**Step 4:** Generate a child solution by applying a crossover operator to the selected parents in the mating pool. A simple one-point crossover operator is used for this step, in which a crossover point is randomly selected, and the child solution will consist of the first j genes taken from the first parent and the remaining (l-j) genes taken from the second parent, or vice versa, with equal probabilities [2]. Figure 11 shows the code snippet and an example of it for the small problem instance.

```
function C=CX_beasly(pop)

parent1 =pop(1,:);
parent2 = pop(2,:);

l = length(parent1);
j=randi([1 l-1]);
C1=[parent1(1:j) parent2(j+1:end)];
C2=[parent2(1:j) parent1(j+1:end)];

r = randi([1, 2], 1);
if r==1
    C=C1;
else
    C=C2;
end

end
```

```
M_pool = 2×6
      3    1    1    2    4    1
      2    3    4    1    1    2

child = 1×6
      3    1    4    1    1    2
```

Figure. 11: step 4: one-point crossover operation

In this example, the second position and the first child are randomly chosen as the crossover point and the final child.

**Step 5:** The crossover procedure is followed by a mutation procedure. This mutation procedure involves exchanging elements in two randomly selected genes. It should be noted that mutation will only happen if two exchanged elements are not the same. There might be a case where no mutation happens. Authors in [2] showed that the GA with only the crossover and mutation operators is effective in producing good-quality solutions. Figure 12 shows the code snippet and an example of it for the small problem instance.

```matlab
function  mutated = mutation_beasly(child)

mutated = child;
nvar = length(child);

j1=randi([1 nvar-1]);
j2=randi([j1+1 nvar]);

nj1=mutated(j1);
nj2=mutated(j2);

mutated(j1)=nj2;
mutated(j2)=nj1;

if child == mutated
    disp('No Mutation happened')
else
    disp('mutation happened')
end
end

child = 1×6
     3     1     4     1     1     2

mutation happened
mutated_child = 1×6
     3     1     1     1     4     2
```

Figure. 12: step 5: pairwise exchange mutation

In this example, mutation happened and positions 3 and 5 were exchanged.

**Step 6:** In this step, we will handle the unfitness of the generated child using the method introduced in [2]. Please note that this step will only happen for unfit children and will be skipped for those with unfitness equal to zero (i.e., no unfitness). For each location in the solution, if the resource capacity of the location is exceeded (i.e., overused location), then a single randomly selected machine is reassigned from the overused location to the next underused location (in order) that has adequate remaining capacity (if one can be found). The result will be a new child with less or no unfitness. Figure 13 shows the code snippet and an example of it for the small problem instance.

```matlab
function [child, fit_child,unfitness_child] = beasly_unfit(child,C,R,F,D,A,T)
[extra_cap_child,unfitness_child] = unfitness_calc(child,C,R);
if unfitness_child >0
    disp('there are unfitness in solutions')
    over_used_loc = find(extra_cap_child(:)>0);
    under_used_loc = find(extra_cap_child(:)<0);
    num_overused = size(over_used_loc,1);
    for k=1:num_overused
        %find all the m/c's that is assigned to the overused loc
        mc = find(child(:) == over_used_loc(k));
        choose = mc(randperm(numel(mc),1)); %Randomly choose one of them
        if ~isempty(under_used_loc)
            extra = min(extra_cap_child(under_used_loc));
            if ~extra<R(choose)
                continue
            else
                inde = find(extra_cap_child==extra,1);
                child(choose) =inde; %assign the m/c to an underused loc
            end
        end
        [extra_cap_child,~] = unfitness_calc(child,C,R);
        under_used_loc = find(extra_cap_child(:)<0);

    end
    fit_child = costcalc_B(child,F,D,A,T);
    [~,unfitness_child]= unfitness_calc(child,C,R);
    disp('The new child is:')
    disp(child)
    disp('the new child unfitness is:')
    disp(unfitness_child)
else
    disp('no unfitness')
    fit_child = costcalc(child,F,D,A,T);
end
end
```

```
mutated_child = 1×6
    3     1     4     1     1     2

there are unfitness in solutions
The new child is:
    3     2     4     1     1     2

the new child unfitness is:
    0
new_child = 1×6
    3     2     4     1     1     2

fit_child = 21265
unfitness_child = 0
```

Figure. 13: step 6: handling unfitness

In this example, there were unfitness in the solution. So, the algorithm changed the assignments somehow and the new child unfitness is zero.

**Step 7:** The child solution will take the place of one chromosome in the population and the new population is generated. In our population replacement strategy, the child replaces the population solution that has the highest unfitness value (i.e., the most unfit solution). If the population consists of all feasible solutions, the solution with the maximum fitness is removed. This replacement plan aids in eliminating infeasible solutions in the population. It should be noted that a duplicate child, which is defined as a solution whose structure is identical to any of the solution structures already in the population, is not permitted to enter the new population because, in that case, the population might end up being made up entirely of identical solutions, which would severely restrict the GA's ability to produce new solutions. Figure 14 shows the code snippet and an example of it for the small problem instance.

```
P1=P0;
check_duplicate = all(ismember(child,P1,'rows')==1);
if check_duplicate
    disp('generated child exist in the solution, duplicate solution')
    continue
else
    %check if there is unfitness in the solution
    check = ~isempty(find(unfitness_P0, 1));
    if check==0
        if unfitness_child==0
            disp('no unfitness remained, exchange the new child with the max fitness')
            [highest_fit, ind] = max(fitness_P0);
            P1(ind,:) = child;
        end
    else
        disp('exchange the new child with max unfitness')
        [highest_unfit, ind] = max(unfitness_P0);
        P1(ind,:) = child;
    end
```

```
The new child is:
    2      3      2      1      1      4

the new child unfitness is:
    20

overall fitness of previous generation is:
    9.7298e+04

overall unfitness of previous generation is:
    150

previous generation is:
        1      2      3      1      4      1   22273      10
        3      1      1      2      4      1   19373      10
        2      3      4      1      1      2   19610      20
        3      1      4      2      1      1   17165       0
        1      1      4      2      4      1   18878     110

exchange the new child with max unfitness
new overall fitness is:
        99015

new overal unfitness is
    60

new generation is:
        1      2      3      1      4      1   22273      10
        3      1      1      2      4      1   19373      10
        2      3      4      1      1      2   19610      20
        3      1      4      2      1      1   17165       0
        2      3      2      1      1      4   20595      20
```

Figure. 14: step 7: update the population

In this example, the newly generated child with unfitness equal to 20 substitutes for the solution with unfitness equal to 110, thus reducing the overall unfitness of the population.

Finally, the best-found solution and best-found objective function value are updated, and the algorithm goes back to step 2. Steps 2 to 7 are repeated until one of the previously mentioned stopping criteria is met. The MATLAB output for a few complete iterations is printed in the appendix to check how the algorithm works.

The GA algorithm is followed by applying a local neighborhood search technique with steepest descend improvement strategy. This make sure that the result of the GA is at local (or hopefully global) optima. If the solution is not at the local optima, the local neighborhood search algorithm will generate a new solution at the local optima. Figure 15 shows the code snippet.

```
improvement = 1;
iter = 0;
while improvement
    iter = iter + 1;
    i = 0;
    NBH_child = steep_desc(Best_sol,C,R);
    cost_NBH = costcalc(NBH_child,F,D,A,T);
    [~,unfit_NBH] = unfitness_calc(NBH_child,C,R);
    final = [NBH_child int64(cost_NBH) unfit_NBH];
    found = find(unfit_NBH==0);
    fitted = final(found,:);
    [mincost, minind] = min(fitted(:,M+1));

    if mincost < Z_best
        disp('a better sol found')
        Best_sol = fitted(minind,1:M);
        disp(Best_sol)
        Z_best = mincost;
        disp(Z_best)
    else
        improvement = 0;
    end
end
disp(Best_sol);
disp(Z_best);
```

Figure. 15: local neighborhood search with steepest descend

## 4. Computational Results

In this part, the results of implementing this algorithm on three different problem instances (i.e., small, medium, and large) are presented. The results are compared with the results from solving the same problems with MPL/CPLEX software and the results presented in [1]. A MacBook Pro 2018 with a 2.3 GHz quad-core Intel Core i5 CPU and 8 GB of RAM was used as the hardware, along with MATLAB R2022b software.

In this project's algorithm, there are two main parameters that can affect the quality of the solution and the time to reach that solution. The first parameter is the number of populations to generate and update throughout the algorithm (i.e., the *n_pop* variable in codes), and the second parameter is the number of iterations that a new non-duplicate child is generated but the best solution was not improved (i.e., the *max_k* variable in codes). A design of experiment approach with two factors and three levels for each factor (low, medium, high) is adopted to determine the best parameters for each problem instance. Thus, each experiment includes nine runs of the algorithm with different sets of parameters. Table 2 shows the different parameter settings that have been tested for each problem instance.

Table 2. algorithm parameters

| Problem Instance | n_pop (low) | n_pop (medium) | n_pop (high) | Max_k (low) | Max_k (medium) | Max_k (high) |
|---|---|---|---|---|---|---|
| Small | 5 | 10 | 15 | 10 | 40 | 70 |
| Medium | 35 | 50 | 75 | 250 | 500 | 700 |
| Large | 50 | 75 | 100 | 300 | 500 | 700 |

Due to the random nature of many steps in the algorithm, each experiment was repeated three times for diversification purposes. The best performing result in terms of quality (i.e., percent deviation from the best-found solution) and computational time is chosen for the respected parameter setup. Totally, 27 experiments have been done for each problem instance. The summary of the results is in table 3 and the details are in the appendix section.

Table 3. Computational result summary

| Problem Instance | n_pop | Max_k | Time (s) | Z_best_GA | Z_best_C | Z_best_found | %D |
|---|---|---|---|---|---|---|---|
| Small | 5 | 70 | 12.48 | 17165 | 17165 | n/a | 0% |
| Medium | 50 | 250 | 578.91 | 1471896 | 1714264 | 1471896 | 0% |
| Large | 100 | 500 | 2148.0701 | 11261034 | 12878101 | 11217503 | 0.39% |

In table 3, n_pop and Max_k are algorithm parameters that gave the best solution in terms of solution quality and computational time. Time is the amount of elapsed time for the algorithm to complete the experiment run in seconds. Z_best_GA is the objective function value result of our algorithm after finishing the GA and local neighborhood search. Z_best_C is the best-found objective function value using MPL/CPLEX software. Z_best_found is the best-found objective function value from [1]. And finally, %D is the percent deviation between our solution and Z_best_found.

The computational result in this project shows that increasing the number of chromosomes in the population doesn't necessarily guarantee better solution. It also shows that the initial randomly generated population may influence the final solution quality.

## 5. Conclusions

In this project, a modified version of the genetic algorithm developed by [2] was presented and applied to three different instances of the GQAP problem. The results, best-found solutions, and objective function values of the GA are compared with the best-found solutions calculated by solving the problem mathematically with MPL/CPLEX software and other metaheuristics. The results show that the developed GA can produce better results more efficiently and in a fraction of the time compared to MPL/CPLEX software results. And the result is interpretable and can be used to solve real-life problems. It should be noted that the main objective of this project was to implement the GA and receive acceptable results. There is a lot of room for improving the algorithm to perform faster and in a more efficient manner. That can be considered in future works.

# References

[1] McKendall, A., & Li, C. (2017). A tabu search heuristic for a generalized quadratic assignment

   problem. Journal of Industrial and Production Engineering, 34(3), 221-231.

[2] Chu, P. C., & Beasley, J. E. (1997). A genetic algorithm for the generalised assignment problem.

Computers & Operations Research, 24(1), 17-23.

[3] IENG 554 lecture notes, Fall 2022

# Appendix

Here are MATLAB output of few complete iterations of the algorithm for the small instance problem.

```
Iteration Number:
     6
The mating pool is:
   3   1   1   2   4   1
   2   3   4   1   1   2
mutation happened
no unfitness
overall fitness of previous generation is:
   1.0053e+05
overall unfitness of previous generation is:
   260
previous generation is:
   1   2   3   1   4   1   22273    10
   3   1   1   2   4   1   19373    10
   2   3   4   1   1   2   19610    20
   4   4   2   1   1   3   20400   110
   1   1   4   2   4   1   18878   110
exchange the new child with max unfitness
new overall fitness is:
   9.7298e+04
new overall unfitness is
   150
new generation is:
   1   2   3   1   4   1   22273    10
   3   1   1   2   4   1   19373    10
   2   3   4   1   1   2   19610    20
   3   1   4   2   1   1   17165     0
   1   1   4   2   4   1   18878   110
new best sol is:
   3   1   4   2   1   1
new Z is:
   17165
Iteration Number:
     7
The mating pool is:
   3   1   4   2   1   1
   2   3   4   1   1   2
mutation happened
there are unfitness in solutions
The new child is:
   2   3   2   1   1   4
```

the new child unfitness is:
  20
overall fitness of previous generation is:
  9.7298e+04
overall unfitness of previous generation is:
  150
previous generation is:

| 1 | 2 | 3 | 1 | 4 | 1 | 22273 | 10 |
|---|---|---|---|---|---|-------|-----|
| 3 | 1 | 1 | 2 | 4 | 1 | 19373 | 10 |
| 2 | 3 | 4 | 1 | 1 | 2 | 19610 | 20 |
| 3 | 1 | 4 | 2 | 1 | 1 | 17165 | 0 |
| 1 | 1 | 4 | 2 | 4 | 1 | 18878 | 110 |

exchange the new child with max unfitness
new overall fitness is:
  99015
new overall unfitness is
  60
new generation is:

| 1 | 2 | 3 | 1 | 4 | 1 | 22273 | 10 |
|---|---|---|---|---|---|-------|-----|
| 3 | 1 | 1 | 2 | 4 | 1 | 19373 | 10 |
| 2 | 3 | 4 | 1 | 1 | 2 | 19610 | 20 |
| 3 | 1 | 4 | 2 | 1 | 1 | 17165 | 0 |
| 2 | 3 | 2 | 1 | 1 | 4 | 20595 | 20 |

new best sol is:
  3  1  4  2  1  1
new Z is:
  17165
Iteration Number:
  8
The mating pool is:
  3   1   4   2   1   1
  3   1   1   2   4   1
No Mutation happened
there are unfitness in solutions
The new child is:
  3   1   1   2   4   1
the new child unfitness is:
  10
overall fitness of previous generation is:
  99015
overall unfitness of previous generation is:
  60
previous generation is:

| 1 | 2 | 3 | 1 | 4 | 1 | 22273 | 10 |
|---|---|---|---|---|---|-------|-----|
| 3 | 1 | 1 | 2 | 4 | 1 | 19373 | 10 |
| 2 | 3 | 4 | 1 | 1 | 2 | 19610 | 20 |
| 3 | 1 | 4 | 2 | 1 | 1 | 17165 | 0 |
| 2 | 3 | 2 | 1 | 1 | 4 | 20595 | 20 |

generated child exist in the solution, duplicate solution

The details of experiments are as follow:

| | n_pop | max_k | Best_sol | Time (s) | Z_best | Optimal solution | Percent deviation | best from mpl/cplex |
|---|---|---|---|---|---|---|---|---|
| **small instance run1** | 5 | 10 | 4 1 1 1 3 4 | 0.39906 | 9,999,999 | 17,165 | n/a | 17,165 |
| | 10 | 10 | 4 3 1 2 2 1 | 0.12692 | 9,999,999 | 17,165 | n/a | 17,165 |
| | 15 | 10 | 3 1 4 2 1 1 | 14.1555 | 17,165 | 17,165 | 0.00% | 17,165 |
| | 5 | 40 | 3 1 4 2 1 1 | 13.7919 | 17,165 | 17,165 | 0.00% | 17,165 |
| | 10 | 40 | 3 1 4 2 1 1 | 15.1179 | 17,165 | 17,165 | 0.00% | 17,165 |
| | 15 | 40 | 3 1 4 2 1 1 | 14.5252 | 17,165 | 17,165 | 0.00% | 17,165 |
| | 5 | 70 | 3 1 4 2 1 1 | 12.4843 | 17,165 | 17,165 | 0.00% | 17,165 |
| | 10 | 70 | 3 1 4 2 1 1 | 13.161 | 17,165 | 17,165 | 0.00% | 17,165 |
| | 15 | 70 | 3 1 4 2 1 1 | 14.0291 | 17,165 | 17,165 | 0.00% | 17,165 |

| | n_pop | max_k | Best_sol | Time (s) | Z_best | Optimal solution | Percent deviation | best from mpl/cplex |
|---|---|---|---|---|---|---|---|---|
| **small instance run2** | 5 | 10 | 3 1 4 2 1 1 | 22.3587 | 17,165 | 17,165 | 0.00% | 17,165 |
| | 10 | 10 | 3 1 4 2 1 1 | 15.5395 | 17,165 | 17,165 | 0.00% | 17,165 |
| | 15 | 10 | 3 1 1 4 1 2 | 0.19471 | 9,999,999 | 17,165 | n/a | 17,165 |
| | 5 | 40 | 3 1 1 4 1 2 | 18.7781 | 17,165 | 17,165 | 0.00% | 17,165 |
| | 10 | 40 | 3 1 1 4 1 2 | 19.0438 | 17,165 | 17,165 | 0.00% | 17,165 |
| | 15 | 40 | 3 1 1 4 1 2 | 9.8093 | 17,165 | 17,165 | 0.00% | 17,165 |
| | 5 | 70 | 3 1 1 4 1 2 | 8.1287 | 17,165 | 17,165 | 0.00% | 17,165 |
| | 10 | 70 | 3 1 1 4 1 2 | 8.543 | 17,165 | 17,165 | 0.00% | 17,165 |
| | 15 | 70 | 3 1 1 4 1 2 | 15.5099 | 17,165 | 17,165 | 0.00% | 17,165 |

| | n_pop | max_k | Best_sol | Time (s) | Z_best | Optimal solution | Percent deviation | best from mpl/cplex |
|---|---|---|---|---|---|---|---|---|
| **small instance run2** | 5 | 10 | 3 1 4 2 1 1 | 19.671 | 17,165 | 17,165 | 0.00% | 17,165 |
| | 10 | 10 | 4 1 1 3 2 2 | 0.071745 | 9,999,999 | 17,165 | n/a | 17,165 |
| | 15 | 10 | 3 1 4 2 1 1 | 18.3361 | 17,165 | 17,165 | 0.00% | 17,165 |
| | 5 | 40 | 3 1 4 2 1 1 | 14.8265 | 17,165 | 17,165 | 0.00% | 17,165 |
| | 10 | 40 | 3 1 4 2 1 1 | 15.0293 | 17,165 | 17,165 | 0.00% | 17,165 |
| | 15 | 40 | 3 1 4 2 1 1 | 14.9304 | 17,165 | 17,165 | 0.00% | 17,165 |
| | 5 | 70 | 3 1 4 2 1 1 | 15.0286 | 17,165 | 17,165 | 0.00% | 17,165 |
| | 10 | 70 | 3 1 4 2 1 1 | 15.4842 | 17,165 | 17,165 | 0.00% | 17,165 |
| | 15 | 70 | 3 1 4 2 1 1 | 14.8255 | 17,165 | 17,165 | 0.00% | 17,165 |

| | n_pop | max_k | Best_sol | Time (s) | Z_best | Z_best after steep | Optimal solution | Percent deviation | best from mpl/cplex |
|---|---|---|---|---|---|---|---|---|---|
| medium instance run1 | 35 | 250 | 1 13 13 9 13 9 13 13 13 3 5 8 5 1 3 | 710.4471 | 1,590,241 | 1,577,883 | 1,471,896 | 7.20% | 1,714,264 |
| | 50 | 250 | 1 13 13 9 9 5 5 13 6 6 5 10 13 9 6 10 5 13 5 1 9 | 885.8834 | 1,640,973 | 1,528,964 | 1,471,896 | 3.88% | 1,714,264 |
| | 75 | 250 | 2 6 10 15 10 5 2 2 6 10 2 6 10 6 10 5 2 5 1 10 | 1074.7386 | 1,471,896 | 1,471,896 | 1,471,896 | 0.00% | 1,714,264 |
| | 35 | 500 | 1 13 13 12 5 5 13 13 1 1 13 1 1 12 9 5 8 5 9 9 | 502.5842 | 1628247 | 1,597,754 | 1,471,896 | 8.55% | 1,714,264 |
| | 50 | 500 | 1 13 13 13 5 13 13 9 13 9 13 9 6 9 5 8 5 1 9 | 596.1952 | 1,517,741 | 1,517,741 | 1,471,896 | 3.11% | 1,714,264 |
| | 75 | 500 | 2 6 10 15 10 5 2 2 6 10 2 6 10 6 10 5 2 5 1 10 | 983.9566 | 1,471,896 | 1,471,896 | 1,471,896 | 0.00% | 1,714,264 |
| | 35 | 700 | 1 13 13 12 5 9 13 6 1 5 5 13 1 6 9 5 13 5 9 3 | 454.9406 | 1,669,290 | 1,594,282 | 1,471,896 | 8.31% | 1,714,264 |
| | 50 | 700 | 2 6 10 1 5 5 2 2 6 10 10 6 10 5 2 5 10 11 | 801.3054 | 1,555,036 | 1,500,497 | 1,471,896 | 1.94% | 1,714,264 |
| | 75 | 700 | 2 6 10 1 5 5 2 6 10 10 10 2 2 5 8 5 6 10 | 1109.7561 | 1,584,395 | 1,584,395 | 1,471,896 | 7.64% | 1,714,264 |
| medium instance run2 | 35 | 250 | 1 9 13 12 13 5 13 13 1 1 11 1 1 12 3 9 13 5 5 9 | 508.9023 | 1,727,122 | 1,681,000 | 1,471,896 | 14.21% | 1,714,264 |
| | 50 | 250 | 2 11 10 12 5 5 2 2 1 10 2 10 10 12 10 10 2 5 1 11 | 598.2446 | 1,728,962 | 1,506,593 | 1,471,896 | 2.36% | 1,714,264 |
| | 75 | 250 | 2 6 10 15 10 5 2 2 6 10 2 6 10 6 10 5 2 5 1 9 | 684.0644 | 1,533,373 | 1,500,497 | 1,471,896 | 1.94% | 1,714,264 |
| | 35 | 500 | 2 6 3 1 10 5 2 2 10 6 10 6 10 5 2 5 10 3 | 569.3753 | 1,623,531 | 1,623,531 | 1,471,896 | 10.30% | 1,714,264 |
| | 50 | 500 | 13 13 13 1 3 9 3 4 13 5 4 13 9 5 10 5 13 5 9 3 | 644.602 | 1,747,772 | 1,654,298 | 1,471,896 | 12.39% | 1,714,264 |
| | 75 | 500 | 2 6 10 15 10 5 2 2 6 10 2 6 10 6 10 5 2 5 1 10 | 741.2125 | 1,471,896 | 1,471,896 | 1,471,896 | 0.00% | 1,714,264 |
| | 35 | 700 | 1 13 9 12 13 5 13 13 1 1 13 13 9 5 9 9 8 5 1 3 | 589.2322 | 1,558,316 | 1,514,852 | 1,471,896 | 2.92% | 1,714,264 |
| | 50 | 700 | 1 13 13 12 13 9 13 6 1 1 5 1 1 6 9 5 8 5 13 9 | 636.1666 | 1,618,967 | 1,568,709 | 1,471,896 | 6.58% | 1,714,264 |
| | 75 | 700 | 2 11 10 12 5 5 2 2 1 10 2 1 1 11 2 5 8 5 10 10 | 647.8656 | 1,734,110 | 1,511,402 | 1,471,896 | 2.68% | 1,714,264 |
| medium instance run3 | 35 | 250 | 1 13 13 15 5 13 6 6 5 4 13 9 6 9 5 13 5 1 9 | 431.5769 | 1,528,964 | 1,528,964 | 1,471,896 | 3.88% | 1,714,264 |
| | 50 | 250 | 2 6 10 15 10 5 2 2 6 10 2 6 10 6 10 5 2 5 1 10 | 578.9126 | 1,471,896 | 1,471,896 | 1,471,896 | 0.00% | 1,714,264 |
| | 75 | 250 | 2 11 10 12 5 5 2 2 1 10 2 1 1 11 2 5 8 5 10 10 | 743.9031 | 1,734,110 | 1,511,402 | 1,471,896 | 2.68% | 1,714,264 |
| | 35 | 500 | 1 13 13 9 15 3 9 13 9 10 13 10 5 10 5 13 5 9 3 | 497.8013 | 1,749,712 | 1,648,016 | 1,471,896 | 11.97% | 1,714,264 |
| | 50 | 500 | 2 6 10 1 5 5 2 2 6 10 2 10 10 6 10 5 2 5 10 11 | 640.9521 | 1,555,036 | 1,500,497 | 1,471,896 | 1.94% | 1,714,264 |
| | 75 | 500 | 1 13 9 15 13 9 13 13 9 13 13 9 5 9 5 8 5 1 9 | 1054.9346 | 1,497,245 | 1,497,245 | 1,471,896 | 1.72% | 1,714,264 |
| | 35 | 700 | 2 11 10 1 5 5 2 2 6 6 10 2 10 10 6 10 5 2 5 10 3 | 559.0019 | 1,618,153 | 1,597,729 | 1,471,896 | 8.55% | 1,714,264 |
| | 50 | 700 | 1 13 9 15 13 9 13 13 9 13 9 5 9 5 8 5 1 9 | 472.4605 | 1,497,245 | 1,497,245 | 1,471,896 | 1.72% | 1,714,264 |
| | 75 | 700 | 1 13 13 15 13 5 13 13 9 13 9 13 9 6 9 5 8 5 1 9 | 941.5869 | 1,517,741 | 1,517,741 | 1,471,896 | 3.11% | 1,714,264 |

| | n_pop | max_k | Best_sol | Time (s) | Z_best | Z_best after steep | Optimal solution | Percent deviation | best from mpl/cplex |
|---|---|---|---|---|---|---|---|---|---|
| large instance run1 | 50 | 500 | 4 7 2 9 4 | 1330.793 | 11,330,290 | 11,330,290 | 11,217,503 | 1.01% | 12,878,101 |
| | 75 | 500 | 6 7 2 2 4 | 1965.7089 | 11,670,235 | 11,670,235 | 11,217,503 | 4.04% | 12,878,101 |
| | 100 | 500 | 4 7 2 9 4 3 | 1707.5355 | 11,719,641 | 11,571,085 | 11,217,503 | 3.15% | 12,878,101 |
| | 50 | 700 | 6 7 2 9 4 | 791.9466 | 11,283,388 | 11,283,388 | 11,217,503 | 0.59% | 12,878,101 |
| | 75 | 700 | 6 7 2 9 4 | 1237.4738 | 11,578,460 | 11,549,256 | 11,217,503 | 2.96% | 12,878,101 |
| | 100 | 700 | 6 7 3 2 1 3 | 1555.2493 | 12,020,532 | 11,672,510 | 11,217,503 | 4.06% | 12,878,101 |
| | 50 | 300 | 6 7 2 3 1 | 1293.3772 | 11,724,203 | 11,724,203 | 11,217,503 | 4.52% | 12,878,101 |
| | 75 | 300 | 6 7 3 9 1 | 2390.7043 | 11,668,947 | 11,668,947 | 11,217,503 | 4.02% | 12,878,101 |
| | 100 | 300 | 4 7 2 9 4 3 | 3278.8803 | 11,878,414 | 11,462,982 | 11,217,503 | 2.19% | 12,878,101 |

| | n_pop | max_k | Best_sol | Time (s) | Z_best | Z_best after steep | Optimal solution | Percent deviation | best from mpl/cplex |
|---|---|---|---|---|---|---|---|---|---|
| large instance run2 | 50 | 500 | 6 7 2 9 4 | 707.1405 | 11,603,906 | 11,603,906 | 11,217,503 | 3.44% | 12,878,101 |
| | 75 | 500 | 5 7 2 9 4 | 1268.8264 | 11,479,174 | 11,471,626 | 11,217,503 | 2.27% | 12,878,101 |
| | 100 | 500 | 4 7 2 9 4 | 2148.0701 | 11,261,034 | 11,261,034 | 11,217,503 | 0.39% | 12,878,101 |
| | 50 | 700 | 4 7 2 9 4 7 | 971.438 | 11,953,262 | 11,566,424 | 11,217,503 | 3.11% | 12,878,101 |
| | 75 | 700 | 4 7 2 9 4 | 1725.4551 | 11,583,934 | 11,555,758 | 11,217,503 | 3.02% | 12,878,101 |
| | 100 | 700 | 6 7 2 9 1 | 2487.14 | 11,606,157 | 11,606,157 | 11,217,503 | 3.46% | 12,878,101 |
| | 50 | 300 | 4 7 2 9 4 | 1432.7039 | 11,294,559 | 11,294,559 | 11,217,503 | 0.69% | 12,878,101 |
| | 75 | 300 | 6 7 2 9 9 | 2664.752 | 11,455,622 | 11,403,051 | 11,217,503 | 1.65% | 12,878,101 |
| | 100 | 300 | 4 7 2 9 4 | 3822.363 | 11,501,651 | 11,401,757 | 11,217,503 | 1.64% | 12,878,101 |

| | n_pop | max_k | Best_sol | Time (s) | Z_best | Z_best after steep | Optimal solution | Percent deviation | best from mpl/cplex |
|---|---|---|---|---|---|---|---|---|---|
| large instance run2 | 50 | 500 | 4 7 2 9 4 3 | 1286.9387 | 11,745,193 | 11547388 | 11,217,503 | 2.94% | 12,878,101 |
| | 75 | 500 | 4 7 2 9 4 3 | 1060.8528 | 11,673,681 | 11,361,707 | 11,217,503 | 1.29% | 12,878,101 |
| | 100 | 500 | 4 7 2 9 9 | 2354.2488 | 11,771,191 | 11,644,161 | 11,217,503 | 3.80% | 12,878,101 |
| | 50 | 700 | 5 7 2 9 9 | 1019.8932 | 11,388,729 | 11,388,729 | 11,217,503 | 1.53% | 12,878,101 |
| | 75 | 700 | 4 7 2 9 4 | 1427.2514 | 11,407,952 | 11,401,757 | 11,217,503 | 1.64% | 12,878,101 |
| | 100 | 700 | 6 7 2 9 4 | 2133.735 | 11,688,436 | 11688436 | 11,217,503 | 4.20% | 12,878,101 |
| | 50 | 300 | 4 7 2 10 4 | 775.6446 | 11,397,978 | 11,397,978 | 11,217,503 | 1.61% | 12,878,101 |
| | 75 | 300 | 4 3 2 4 4 3 | 2162.0918 | 11,958,708 | 11,545,493 | 11,217,503 | 2.92% | 12,878,101 |
| | 100 | 300 | 4 7 2 9 4 | 2625.6873 | 11,657,609 | 11,644,121 | 11,217,503 | 3.80% | 12,878,101 |

Figure. 16: computational results detail